\documentclass[11pt,letterpaper]{article}
\usepackage{emnlp2017}
\usepackage{times}
\usepackage{latexsym}
\usepackage{amsmath}
\usepackage{booktabs}
\usepackage[utf8]{inputenc}
\usepackage[boxed]{algorithm2e}
\usepackage{listings}
\usepackage{float}
\usepackage{graphicx}
\usepackage[compact]{titlesec}

\emnlpfinalcopy

\def\emnlppaperid{***}
\title{End-to-End Information Extraction without Token-Level Supervision}

\author{
	Rasmus Berg Palm \\ DTU Compute \\ Technical University of Denmark \\ \texttt{rapal@dtu.dk}
    \And
    Dirk Hovy \\ Computer Science Dpeartment \\ University of Copenhagen  \\ \texttt{dirk.hovy@di.ku.dk}
    \AND
    Florian Laws \\ Tradeshift \\ Landemærket 10, 1119 Copenhagen \\ \texttt{fla@tradeshift.com}
    \And
    Ole Winther \\ DTU Compute \\ Technical University of Denmark \\ \texttt{olwi@dtu.dk}
}

\date{}

\begin{document}

\maketitle

\begin{abstract}
Most state-of-the-art information extraction approaches rely on token-level labels to find the areas of interest in text. Unfortunately, these labels are time-consuming and costly to create, and consequently, not available for many real-life IE tasks. To make matters worse, token-level labels are usually not the desired output, but just an intermediary step. 
End-to-end (E2E) models, which take raw text as input and produce the desired output directly, need not depend on token-level labels. 
We propose an E2E model based on pointer networks, which can be trained directly on pairs of raw input and output text.
We evaluate our model on the ATIS data set, MIT restaurant corpus and the MIT movie corpus and compare to neural baselines that do use token-level labels. We achieve competitive results, within a few percentage points of the baselines, showing the feasibility of E2E information extraction without the need for token-level labels.
This opens up new possibilities, as for many tasks currently addressed by human extractors, raw input and output data are available, but not token-level labels.
\end{abstract}

\begin{figure*}[t]
	\centering \includegraphics[width=1.0\textwidth]{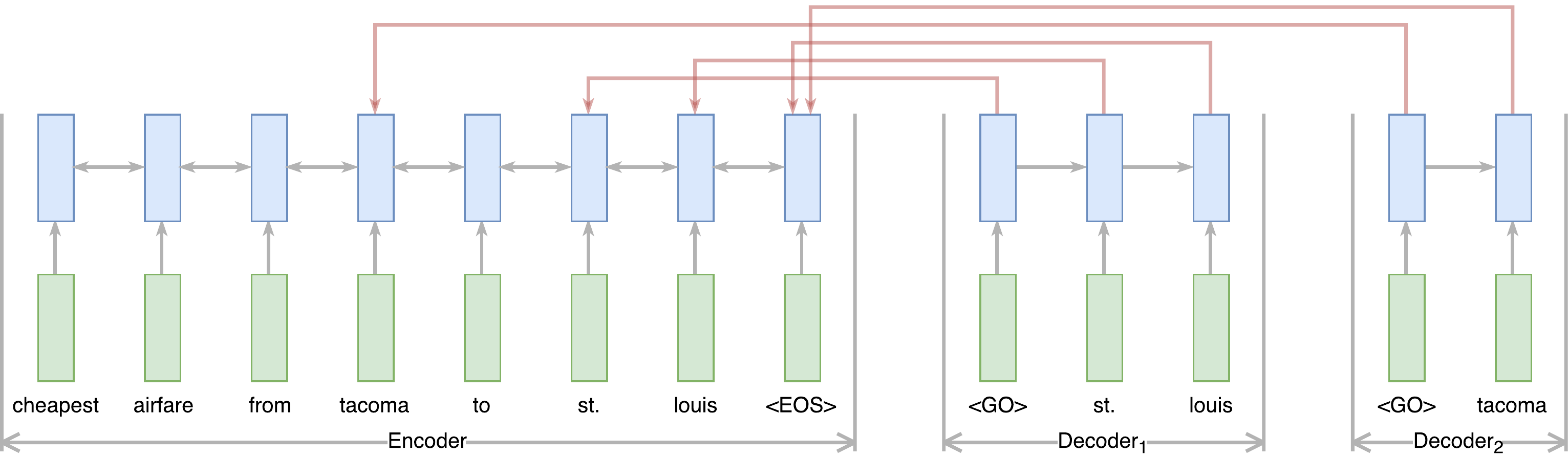}
    \caption{Our model based on pointer networks. The solid red lines are the attention weights. For clarity only two decoders are drawn and only the strongest attention weight for each output is drawn.}
    \label{fig:model}
\end{figure*}

\section{Introduction}
Humans spend countless hours extracting structured machine readable information from unstructured information in a multitude of domains. Promising to automate this, information extraction (IE) is one of the most sought-after industrial applications of natural language processing. However, despite substantial research efforts, in practice, many applications still rely on manual effort to extract the relevant information.

One of the main bottlenecks is a shortage of the data required to train state-of-the-art IE models, which rely on sequence tagging \cite{finkel_incorporating_2005,zhai_neural_2017}. Such models require sufficient amounts of training data that is labeled at the token-level, i.e., with one label for each word.

The reason token-level labels are in short supply is that they are not the intended output of human IE tasks. Creating token-level labels thus requires an additional effort, essentially doubling the work required to process each item. This additional effort is expensive and infeasible for many production systems: estimates put the average cost for a sentence at about 3 dollars, and about half an hour annotator time \cite{alonso_noisy_2016}. Consequently, state-of-the-art IE approaches, relying on sequence taggers, cannot be applied to many real life IE tasks.

What is readily available in abundance and at no additional costs, is the raw, unstructured input and machine-readable output to a human IE task. Consider the transcription of receipts, checks, or business documents, where the input is an unstructured PDF and the output a row in a database (due date, payable amount, etc). Another example is flight bookings, where the input is a natural language request from the user, and the output a HTTP request, sent to the airline booking API.

To better exploit such existing data sources, we propose an end-to-end (E2E) model based on pointer networks with attention, which can be trained end-to-end on the input/output pairs of human IE tasks, without requiring token-level annotations.

We evaluate our model on three traditional IE data sets. Note that our model and the baselines are competing in two dimensions. The first is cost and applicability. The baselines require token-level labels, which are expensive and unavailable for many real life tasks. Our model does \textit{not} require such token-level labels. Given the time and money required for these annotations, our model clearly improves over the baselines in this dimension.
The second dimension is the accuracy of the models. Here we show that our model is competitive with the baseline models on two of the data sets and only slightly worse on the last data set, all despite fewer available annotations.

\paragraph{Contributions}
We present an E2E IE model with attention that does not depend on costly token-level labels, yet performs competitively with neural baseline models that rely on token-level labels. By saving both time and money at comparable performance, our model presents a viable alternative for many real-life IE needs. Code is available at \href{https://github.com/rasmusbergpalm/e2e-ie-release}{github.com/rasmusbergpalm/e2e-ie-release}

\section{Model}
Our proposed model is based on pointer networks \cite{vinyals_pointer_2015}. A pointer network is a sequence-to-sequence model with attention in which the output is a position in the input sequence. The input position is "pointed to" using the attention mechanism. See figure \ref{fig:model} for an overview. Our formulation of the pointer network is slightly different from the original: Our output is some content from the input rather than a position in the input.

An input sequence of $N$ words $\mathbf{x} = x_1,...,x_N$ is encoded into another sequence of length $N$ using an Encoder.
\begin{align}
e_i &= \text{Encoder}(x_i, e_{i-1})
\end{align}
We use a single shared encoder, and $k = 1..K$ decoders, one for each piece of information we wish to extract. At each step $j$ each decoder calculate an unnormalized scalar attention score $a_{kji}$ over each input position $i$. The $k$'th decoder output at step $j$, $o_{kj}$, is then the weighted sum of inputs, weighted with the normalized attention scores $att_{kji}$.
\begin{align}
d_{kj} &= \text{Decoder}_k(o_{k,j-1}, d_{k,j-1}) \\
a_{kji} &= \text{Attention}_k(d_{kj}, e_i) \text{ for } i = 1..N \\
att_{kji} &= \text{softmax}(a_{kji}) \text{ for } i = 1..N \\
o_{kj} &= \sum_{i=1}^N att_{kji} \, x_i \ . 
\end{align}
Since each $x_i$ is a one-hot encoded word, and the $att_{kji}$ sum to one over $i$, $o_{kj}$ is a probability distribution over words.

The loss function is the sum of the negative cross entropy for each of the expected outputs $y_{kj}$ and decoder outputs $o_{kj}$.
\begin{align}
	\mathcal{L}(\bf{x}, \bf{y}) &= -\sum_{k=1}^K \frac{1}{M_k} \sum_{j=1}^{M_k}  y_{kj} \log \left(o_{kj}\right) \ ,
\end{align}
where $M_k$ is the sequence length of expected output $y_k$.

The specific architecture depends on the choice of $\text{Encoder}$, $\text{Decoder}$ and $\text{Attention}$. For the encoder, we use a Bi-LSTM with 128 hidden units and a word embedding of 96 dimensions. We use a separate decoder for each of the fields. Each decoder has a word embedding of 96 dimensions, a LSTM with 128 units, with a learned first hidden state and its own attention mechanism. Our attention mechanism follows \citet{bahdanau_neural_2014}
\begin{align}
	a_{ji} &= v^T \tanh(W_{e} \, enc_i + W_{d} \, dec_j) \ .
\end{align}
The attention parameters $W_e$, $W_d$ and $v$ for each attention mechanism are all 128-dimensional. 

During training we use teacher forcing for the decoders \cite{williams_learning_1989}, such that $o_{k,j-1} = y_{k,j-1}$. During testing we use argmax to select the most probable output for each step $j$ and run each decoder until the first end of sentence (EOS) symbol.

\section{Experiments}
\subsection{Data sets}
To compare our model to baselines relying on token-level labels we use existing data sets for which token level-labels are available. We measure our performance on the ATIS data set \cite{price_evaluation_1990} (4978 training samples, 893 testing samples) and the MIT restaurant (7660 train, 1521 test) and movie corpus (9775 train, 2443 test) \cite{liu_asgard:_2013}. These data sets contains token-level labels in the Beginning-Inside-Out format (BIO).

The ATIS data set consists of natural language requests to a simulated airline booking system. Each word is labeled with one of several classes, e.g. departure city, arrival city, cost, etc. The MIT restaurant and movie corpus are similar, except for a restaurant and movie domain respectively. See table \ref{table:mit-samples} for samples.

\begin{table}[!ht]
  \begin{centering}
  \begin{tabular}{lrlr}
  \multicolumn{2}{c}{\bf{MIT Restaurant}} & \multicolumn{2}{c}{\bf{MIT Movie}} \\
2 & \texttt{B-Rating} & show & \texttt{O} \\
start & \texttt{I-Rating} & me & \texttt{O} \\
restaurants & \texttt{O} & films & \texttt{O} \\
with & \texttt{O} & elvis & \texttt{B-ACTOR}  \\
inside & \texttt{B-Amenity} & films & \texttt{O}  \\
dining & \texttt{I-Amenity} & set & \texttt{B-PLOT} \\
& & in & \texttt{I-PLOT} \\
& & hawaii& \texttt{I-PLOT}
  \end{tabular}
  \caption{Samples from the MIT restaurant and movie corpus. The transcription errors are part of the data. \label{table:mit-samples}}
  \end{centering}
\end{table}

Since our model does not need token-level labels, we create an E2E version of each data set without token-level labels by chunking the BIO-labeled words and using the labels as fields to extract. If there are multiple outputs for a single field, e.g. multiple destination cities, we join them with a comma. For the ATIS data set, we choose the 10 most common labels, and we use all the labels for the movie and restaurant corpus. The movie data set has 12 fields and the restaurant has 8. See Table \ref{table:atis-data} for an example of the E2E ATIS data set.

\begin{table}[!ht]
  \begin{centering}
  \begin{tabular}{ll}
  	\multicolumn{2}{c}{\bf{Input}} \\    
    \multicolumn{2}{l}{cheapest airfare from tacoma to st. louis and detroit} \\ \\
    \multicolumn{2}{c}{\bf{Output}} \\
    \texttt{fromloc} & tacoma \\
    \texttt{toloc} & st. louis , detroit \\
    \texttt{airline\_name} & - \\
    \texttt{cost\_relative} & cheapest \\
    \texttt{period\_of\_day} & - \\ 
    \texttt{time} & - \\
    \texttt{time\_relative} & - \\
    \texttt{day\_name} & - \\
    \texttt{day\_number} & - \\
    \texttt{month\_name} & - \\
  \end{tabular}
  \caption{Sample from the E2E ATIS data set. \label{table:atis-data}}
  \end{centering}
\end{table}

\subsection{Baselines}
For the baselines, we use a two layer neural network model. The first layer is a Bi-directional Long Short Term Memory network \cite{hochreiter_long_1997} (Bi-LSTM) and the second layer is a forward-only LSTM. Both layers have 128 hidden units. We use a trained word embedding of size 128. The baseline is trained with Adam \cite{kingma_adam:_2014} on the BIO labels and uses early stopping on a held out validation set.

This baseline architecture achieves a fairly strong F1 score of 0.9456 on the ATIS data set. For comparison, the published state-of-the-art is at 0.9586 \cite{zhai_neural_2017}. These numbers are for the traditional BIO token level measure of performance using the publicly available conlleval script. They should not be confused with the E2E performance reported later. We present them here so that readers familiar with the ATIS data set can evaluate the strength of our baselines using a well-known measure.

For the E2E performance measure we train the baseline models using token-level BIO labels and predict BIO labels on the test set. Given the predicted BIO labels, we create the E2E output for the baseline models in the same way we created the E2E data sets, i.e. by chunking and extracting labels as fields. We evaluate our model and the baselines using the MUC-5 definitions of precision, recall and F1, without partial matches \cite{chinchor_muc-5_1993}. We use bootstrap sampling to estimate the probability that the model with the best micro average F1 score on the entire test set is worse for a randomly sampled subset of the test data.

\subsection{Our model}
Since our decoders can only output values that are present in the input, we prepend a single comma to every input sequence. We optimize our model using Adam and use early stopping on a held-out validation set. The model quickly converges to optimal performance, usually after around 5000 updates after which it starts overfitting.

For the restaurant data set, to increase performance, we double the sizes of all the parameters and use embedding and recurrent dropout following \cite{gal_theoretically_2015}. Further, we add a summarizer LSTM to each decoder. The summarizer LSTM reads the entire encoded input. The last hidden state of the summarizer LSTM is then concatenated to each input to the decoder.

\subsection{Results}

\begin{table}[!ht]
	\setlength{\tabcolsep}{5pt}
	\renewcommand{\arraystretch}{1.3} 
    \begin{centering}
	\begin{tabular}{lccc}
        Data set		& Baseline			& Ours 			& $p$ \\ 
        \midrule
		ATIS			& 0.977				& 0.974 		& 0.1755 \\        
        Movie			& 0.816				& 0.817			& 0.3792 \\
        Restaurant		& \textbf{0.724}	& 0.694 		& 0.0001 \\
    \end{tabular}
    \caption{Micro average F1 scores on the E2E data sets. Results that are significantly better ($p<0.05$) are highlighted in bold.\label{table:results}}
    \end{centering}
\end{table}

We see in Table \ref{table:results} that our model is competitive with the baseline models in terms of micro-averaged F1 for two of the three data sets. This is a remarkable result given that the baselines are trained on token-level labels, whereas our model is trained end-to-end. For the restaurant data set, our model is slightly worse than the baseline.

\section{Related work}
Event extraction (EE) is similar to the E2E IE task we propose, except that it can have several event types and multiple events per input. In our E2E IE task, we only have a single event type and assume there is zero or one event mentioned in the input, which is an easier task. Recently, \citet{nguyen_joint_2016} achieved state of the art results on the ACE 2005 EE data set using a recurrent neural network to jointly model event triggers and argument roles.

Other approaches have addressed the need for token-level labels when only raw output values are available. \newcite{mintz_distant_2009} introduced distant supervision, which heuristically generates the token-level labels from the output values. You do this by searching for input tokens that matches output values. The matching tokens are then assigned the labels for the matching outputs. One drawback is that the quality of the labels crucially depend on the search algorithm and how closely the tokens match the output values, which makes it brittle. Our method is trained end-to-end, thus not relying on brittle heuristics.

\newcite{sutskever_sequence_2014} opened up the sequence-to-sequence paradigm. With the addition of attention \cite{bahdanau_neural_2014}, these models achieved state-of-the-art results in machine translation \cite{wu_googles_2016}. We are broadly inspired by these results to investigate E2E models for IE. 

The idea of copying words from the input to the output have been used in machine translation to overcome problems with out-of-vocabulary words \cite{gulcehre_pointing_2016,gu_incorporating_2016}.

\section{Discussion}

We present an end-to-end IE model that does not require detailed token-level labels. Despite being trained end-to-end, it is competitive with baseline models relying on token-level labels. In contrast to them, our model can be used on many real life IE tasks where intermediate token-level labels are not available and creating them is not feasible.

In our experiments our model and the baselines had access to the same amount of training samples. In a real life scenario it is likely that token-level labels only exist for a subset of all the data. It would be interesting to investigate the quantity/quality trade-of of the labels, and a multi task extension to the model, which could make use of available token-level labels.

Our model is remarkably stable to hyper parameter changes. On the restaurant dataset we tried several different architectures and hyper parameters before settling on the reported one. The difference between the worst and the best was approximately 2 percentage points.

A major limitation of the proposed model is that it can only output values that are present in the input. This is a problem for outputs that are normalized before being submitted as machine readable data, which is a common occurrence. For instance, dates might appear as '\texttt{Jan 17 2012}' in the input and as '\texttt{17-01-2012}' in the machine readable output.

While it is clear that this model does not solve all the problems present in real-life IE tasks, we believe it is an important step towards applicable E2E IE systems. 

In the future, we will experiment with adding character level models on top of the pointer network outputs so the model can focus on an input, and then normalize it to fit the normalized outputs.

\section*{Acknowledgments}
We would like to thank the reviewers who helped make the paper more concise. Dirk Hovy was supported by the Eurostars grant E10138 ReProsis. This research was supported by the NVIDIA Corporation with the donation of TITAN X GPUs.

\bibliographystyle{emnlp_natbib}
\bibliography{Zotero}
\end{document}